\documentclass{article}
\usepackage{graphicx} 
\usepackage[margin=2cm]{geometry}
\usepackage{algorithm}
\usepackage{algpseudocode}
\usepackage{url}
\usepackage{subcaption}
\usepackage{amsfonts}

\title{Contrastive learning for video reuse detection}
\author{Arjun Biswas, Tristan J.B. Cann}
\date{}

\begin{document}

\maketitle

\begin{abstract}
    \noindent The incorporation of footage from other sources has long been a concern from a copyright perspective, as such usage can harm the interests of rights holders. Recent trends in the ``remixing'' of social media content and the development of bespoke stock libraries for specialist causes mean that methods to identify content reuse will have a range of applications. In this paper, we develop a framework based on a Siamese network fine tuned using contrastive learning to identify potential sources for key frames in a given video. This framework makes two contributions to analysis of large video datasets. Firstly, we demonstrate its utility at reducing complex video data to a set of key visuals typically less than 10\% of the size of the full set. Secondly, we show that these reduced sets allow for efficient and effective identification of component footage. We test this method over a set of videos compiled from a stock footage library and demonstrate its effectiveness for constituent video identification with average accuracy greater than 0.95.
\end{abstract}

\section{Introduction}

As access to technology and the quality of internet coverage have increased, so too has the importance of video as a key pillar of social interaction. This ubiquity of cameras and the affordances of new social media platforms mean that more photos and videos are created, shared and reach ever larger sections of society than ever before (e.g. 90\% of US teenagers reported using YouTube in 2024~\cite{Pew_social_media_2024}). In this space of explosive content growth it is becoming close to impossible to track distribution and trends related to specific footage - YouTube reports on average 20 million videos are uploaded to their platform every day\footnote{\url{https://blog.youtube/press/}, accessed 2025-05-12}.

Republishing and reuse of video content is a growing trend, with the potential for wider societal impact~\cite{cnn_video_clippers}. There are a number of legitimate and illegitimate reasons a content producer may choose to include externally-sourced footage in their videos. Illegitimate uses include attempts to share (and potentially profit from) the copyrighted works of others, which does not always include presenting themself as the original creator of the given work. This is a significant concern to many of the largest video hosting platforms who deploy bespoke systems of varying quality to monitor uploaded footage (e.g. the YouTube Content ID system\footnote{\url{https://support.google.com/youtube/answer/2797370?hl=en-GB}, accessed 2026-07-24}). Some of the many legitimate reasons for content reuse reflect shifts in the culture of social media users. Remixing and reaction content have become increasingly popular genres across many platforms, and are even facilitated by platform features (e.g. TikTok ``stitching'' of public videos\footnote{\url{https://support.tiktok.com/en/using-tiktok/creating-videos/stitch-settings}, accessed 2026-07-24}, with implications for how content resonates with audiences~\cite{Bach2023}). In these cases, the inclusion of memes and tropes in novel videos can have significant impacts on the viral success of posts and typically fall within ``fair use'' protections under copyright legislation (although may be contested in certain ways~\cite{Galea2024}). The use cases highlighted above are distinctly focused on the interests of the organisational players in this space - for the rights holders to protect their interests and for the platforms to ensure compliance with legal frameworks. As a result, they are not always able to support the interests of the those beyond these organisations and therefore there is a space for models that can support automated detection of content reuse at intermediate scales.

There is a long history of efforts to measure content similarity. These methods developed alongside the rise in online video sharing platforms such as YouTube in the early 2000s, and considered a wide range of potential techniques for measuring the alignment between two pieces of content. A comparison of the state-of-the-art at this time in~\cite{LawTo07} demonstrates this divergence. The work of Chen et al.~\cite{chen2021efficientobjectembeddingspliced} is of particular relevance in the context of content repetition and remixing given their consideration of spliced image retrieval, i.e. recognising parts of images in addition to the whole. Patch-based approaches such as theirs are a key component of recognition of picture-in-picture and other overlay methods for editing videos. More recently, trends in research have moved away from the more general copy detection problem and towards solutions to more specific societal challenges such as deepfake detection~\cite{Yu2021,Yan_2025_CVPR}

One of the major challenges of working with videos in computer vision is the frequency of redundant visual information. Specifically, if we consider each video as a sequence of frames, then many of the frames will necessarily be minor variations of one another. Many methods for extracting key frames in videos have been employed in the past from those as simple as uniform sampling of frames to those based on scene boundaries~\cite{Calic02} or frame-to-frame variance~\cite{Yoon_2023_CVPR}. Incorporating such steps has clear benefits for model efficiency, and has also been shown to have limited impact on model performance~\cite{Yoon_2023_CVPR}.

Past applications of video downsampling are often aligned with the concept of video fingerprinting, that is the process of extracting some unique representation of a video. A recent literature review in this area by Allouche and Mitrea~\cite{Allouche22} presents a comprehensive view of how the methods for this task have evolved in recent years.  They recognise a split into two camps based on whether a neural network is used, but many common strategies appear in the methods they compare. Other approaches consider a range of temporal features (e.g. the sequence of scene transitions), or local features (e.g. points of interest in key frames) to summarise a given video for subsequent comparison~\cite{7935507,babenko2014neuralcodesimageretrieval}. These are only a small sample of the vast range of possible features for video fingerprinting.

The alignment of meaning in videos is one area that can be overlooked by computer vision methods that seek to only detect matches between (near) identical content. In some applications however, this distinction is not necessary, for example the automated detection of videos conveying similar messages. This consideration has been discussed in the past, primarily in the context of text-to-image retrieval. One such example from Wray et al.~\cite{Wray_2021_CVPR} refers to the semantic alignment in this case and shows how it is an effective tool for identifying image-caption pairs. But this consideration of semantic alignment does not cover all of the features of visual meaning captured by semiotic alignment (e.g. emotional framing). As a result, applications of these computer vision methods should be conscious of whether they seek repetition of specific visuals or the meaning imparted to inform their choice of approach.

Contrastive learning is an important tool in the case of video copy detection, given its utility as a method for distinguishing between similar and dissimilar items. Chen et al.~\cite{Chen20} made a seminal contribution in this area, illustrating the value of self-supervised training and careful data augmentation to provide exemplary performance. Applications of contrastive learning to video data are another challenge however, as noted by Kuang et al.~\cite{Kuang_2021_ICCV}. Their work detailed a means of efficiently generating the positive and negative samples from video data while still capturing the temporal aspects of the video stream. Moreover, they showed state-of-the-art performance on video retrieval over video datasets. Contrastive learning has also shown promise for video highlight detection, that is clips within a video that help distinguish it from others~\cite{Badamdorj_2022_CVPR}.

In this manuscript we present a method based on a Siamese network fine-tuned using contrastive learning to detect reuse of video footage, under the typical conditions of online video sharing. The Siamese network is used in two ways. Firstly, we use the network to produce an analogue of fingerprinting methods by sampling a subset of visually-representative frames that are collectively similar to all frames in a given video. We then use the same Siamese network to compare the representative frames between two or more videos. 

We test this approach on a series of videos constructed by concatenating videos taken from the stock footage library Pixabay and find that our framework for representative frame extraction and subsequent matching is highly effective at both reducing the number of frames to be compared and recovering the included stock footage edited together to form a longer video. When applied using a threshold suitable for our test dataset, we found that on average the representative frame set was approximately 10\% of the frames in a video. During the matching step, we found that our method demonstrates a preference for recall over precision while still achieving an average balanced accuracy of over 0.95.

This paper proceeds as follows. In Section~\ref{sec:methods} we describe the contrastive learning process for our Siamese network, the training data used and outline the process for identifying repeated visuals in new videos, including our creation of a synthetic dataset. In Section~\ref{sec:results} we apply our new method to the synthetic dataset and report the accuracy we observed in identifying the repeated content. Finally, we conclude in Section~\ref{sec:discussion}, and review the implications of our work for further applications of contrastive learning in this space.

\section{Methods}\label{sec:methods}

\begin{figure}[h]
    \centering
    \includegraphics[width=\linewidth]{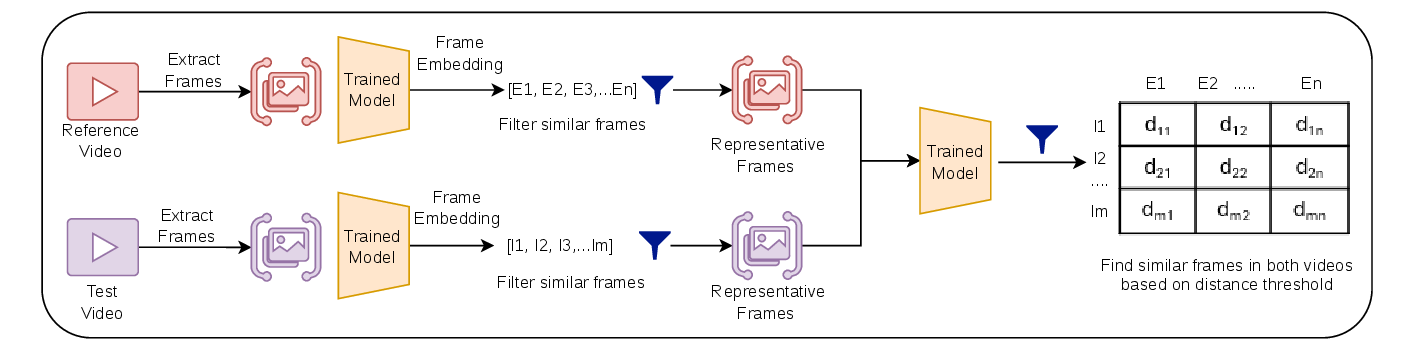}
    \caption{Overall approach for the comparison of videos. We use our trained model to extract a set of representative frames from both the reference and test videos, followed by using the same model to identify similar frames between both videos.}
    \label{fig:overall-approach}
\end{figure}

In this section we describe the constituent steps that form our method for identifying the inclusion of specific portions of footage in longer video data. The key concept in our model is the application of a Siamese network trained using contrastive learning to determine the level of similarity between pairs of frames taken from videos. A schematic of the full process is included in Fig.~\ref{fig:overall-approach}. In short, we use the similarity model in two ways: to efficiently reduce the set of all frames to those that are visually distinct (Section~\ref{sec:downsampling}) and then use these to identify repeated footage (Section~\ref{sec:identifying_reused_footage}). The remainder of this section will describe our training data, Siamese network development and the full details for applying and testing our methodology.

\subsection{Data}\label{sec:data}

Our training data was sampled from ImageNet100~\cite{5206848} given our preference to train a lightweight model usable on consumer-level hardware. A sample of 400 images was randomly selected from every 5th class from ImageNet100 (20 classes in total). This sampling strategy is designed to cover a range of image types for a general purpose model.

In order to construct a video dataset where the included footage is known, we collected videos from Pixabay, a royalty-free stock image and video library. Here users can upload videos in a variety of formats and lengths for reuse by others in their work. All videos we collected were used for evaluation purposes only, and were not included in the training process for our models. A diverse range of videos were collected from Pixabay using five search terms (\textit{people}, \textit{city}, \textit{landscape}, \textit{climate}, \textit{wildlife}). For each of these search terms, we collected the first 25 listed videos after sorting by ``Editor's Choice'' that had a duration of at least 10 seconds. There were rare instances of overlap between the search terms, in such cases additional videos were collected until we were left with a final sample of 125 unique videos. Video URLs are provided in Table~\ref{tab:pixabay urls}. We also ensured that we did not collect any footage on Pixabay produced by generative AI models (as declared by the uploader). While there are no known features that bias our method towards success or failure on such data, we prefer to avoid any potential confounding factors.

\subsection{Siamese network construction}

\begin{figure}
    \centering
    \includegraphics[width=\linewidth]{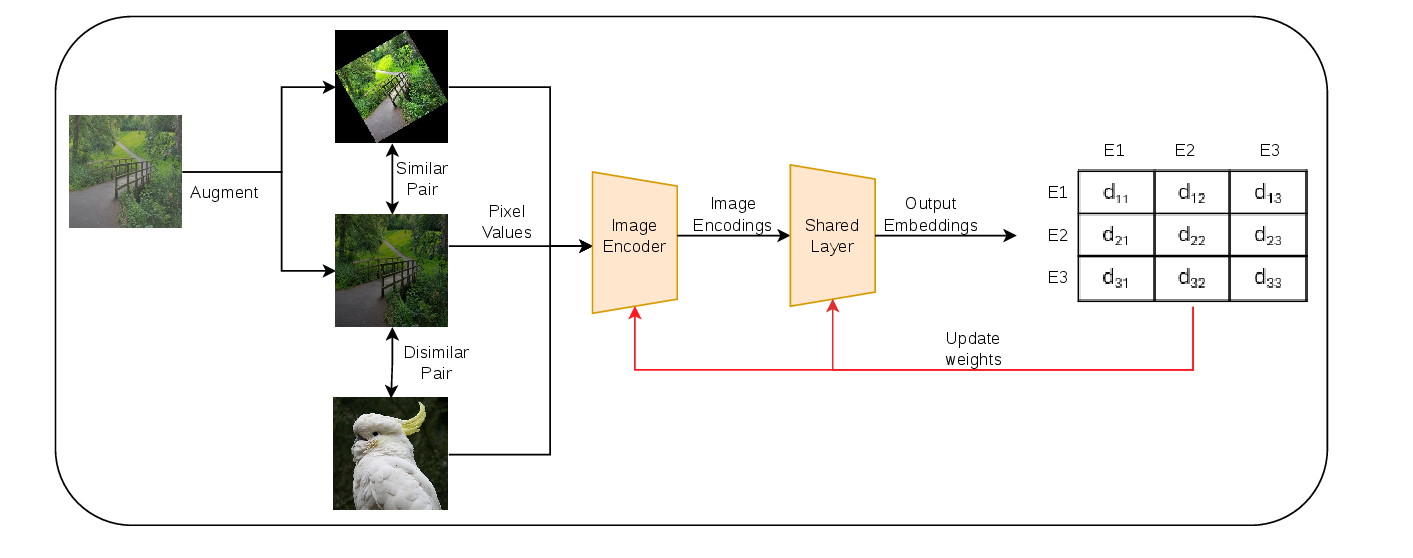}
    \caption{Overview of our contrastive learning pipeline. Input images are augmented to generate positive (similar) pairs, while different images are paired to form negative (dissimilar) pairs. Each image is processed by a shared image encoder followed by a shared fully connected projection layer to produce embedding representations, which are then used for contrastive learning. The bridge image was taken by the authors, the parrot image is taken from ImageNet100~\cite{5206848}.}
    \label{fig:Train-approach}
\end{figure}

Our training pipeline is shown in Fig.~\ref{fig:Train-approach}. At a conceptual level, our Siamese network is trained to compare two images and determine the level of similarity between them. To do this, it returns a distance score, with smaller values indicating greater similarity. We can then use these scores to perform and rank pairwise comparisons of frames in videos (subsequently considered as streams of images for all purposes).

Our model uses a pre-trained backbone to encode a batch of image pairs. The final hidden state of the backbone is used to encode the images, followed by global average pooling. Finally, a fully-connected layer is used to project a pair of images into a shared embedding space. We use an embedding distance-based loss function proposed by Hadsell et al.~\cite{Hadsell2006DimensionalityRB} where pairwise Euclidean distances are calculated between the image embeddings and the loss function is minimised (maximised) for image pairs which are the same (different). We compare the performance of three candidate backbones: ResNet-50~\cite{he2015deepresiduallearningimage}, DenseNet-121~\cite{densenet121} and CLIP-ViT-B-16~\cite{radford2021learningtransferablevisualmodels}. Each trained model uses the same training data of 24,000 image pairs, the same evaluation framework and are included to indicate whether the choice of backbone impacts model performance.

Our training data is divided into two classes: similar and dissimilar pairs. The similar pairs constitute an image paired with a copy of itself, whereas the dissimilar pairs comprise images sampled from two different ImageNet100 classes. In total, we sampled 12,000 similar pairs and 12,000 dissimilar pairs from the dataset described in Section~\ref{sec:data}. 

\begin{figure}
    \centering
    \begin{subfigure}{0.16\textwidth}
        \includegraphics[width=\linewidth]{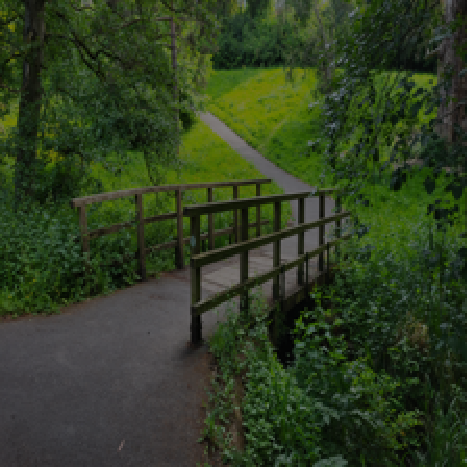}
        \caption{Brightness}
    \end{subfigure}
    \begin{subfigure}{0.16\textwidth}
        \includegraphics[width=\linewidth]{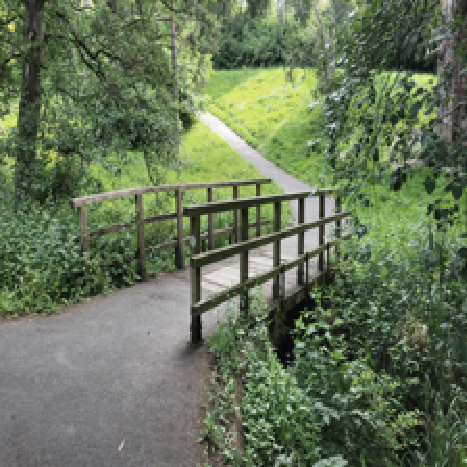}
        \caption{Saturation}
    \end{subfigure}
    \begin{subfigure}{0.16\textwidth}
        \includegraphics[width=\linewidth]{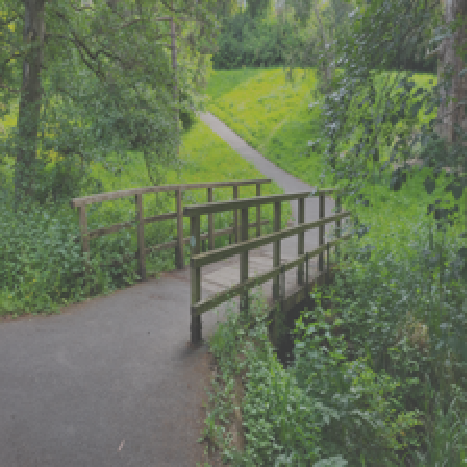}
        \caption{Contrast}
    \end{subfigure}
    \begin{subfigure}{0.16\textwidth}
        \includegraphics[width=\linewidth]{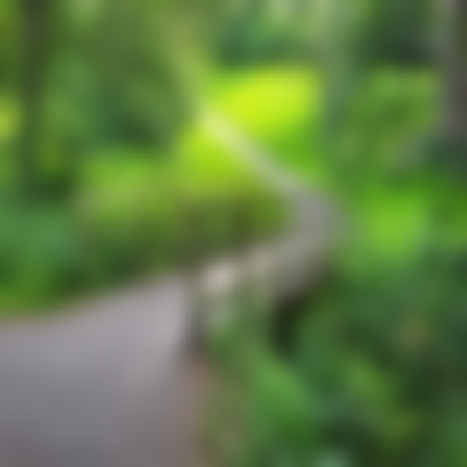}
        \caption{Gaussian blur}
    \end{subfigure}
    \begin{subfigure}{0.16\textwidth}
        \includegraphics[width=\linewidth]{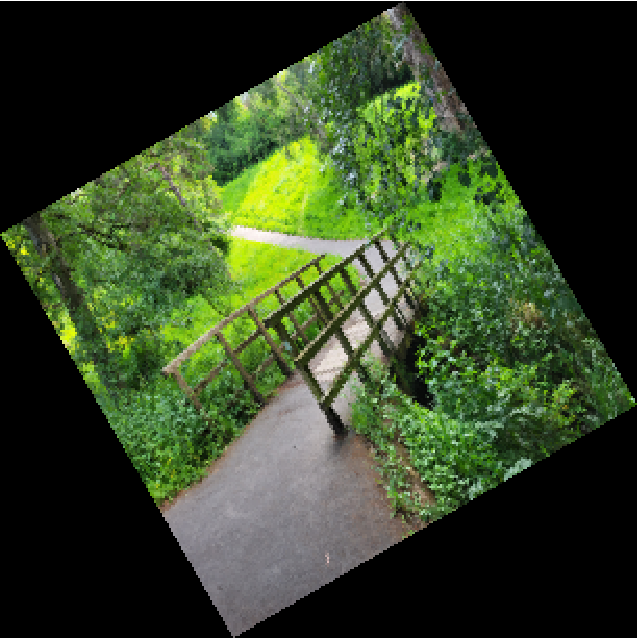}
        \caption{Rotation}
    \end{subfigure}
    \begin{subfigure}{0.16\textwidth}
        \includegraphics[width=\linewidth]{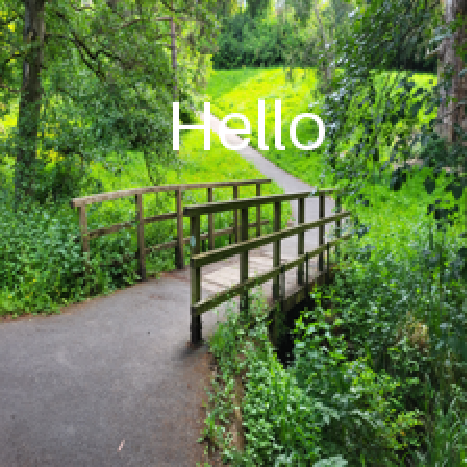}
        \caption{Text overlay}
    \end{subfigure}
    \caption{Examples of the augmentations applied to each the relevant images. The raw image was taken by the authors.}
    \label{fig:augmentations}
\end{figure}

In many of the real-world cases where footage is reused, various editing techniques can be applied. For example, adjustments may be made to colour balance, footage may be cropped or resized to fit with other on-screen elements, or overlays added in the form of text or other imagery. Since we want our model to recognise the underlying footage in such cases, we apply these augmentations to a random sample of image pairs in our training data. We randomly sample 20\% of the data and apply augmentations to one of the images in a pair to simulate certain editing methods. We adjust brightness, saturation or sharpness, or apply Gaussian blur, rotation or text augmentation to a chosen image. For brightness, saturation, and contrast, the augmentation factor is randomly sampled from a normal distribution with a mean of 1.5 and a standard deviation of 0.5. Gaussian blur is applied using a randomly selected kernel size of 5, 7, or 9. Rotation angles are randomly sampled from the range of -30 degrees to 30 degrees. Finally, the text ``Hello world'' is overlaid in white onto the selected image. Each augmentation was applied independently to the randomly selected image pairs, with a single augmentation operation sampled and applied to one image in each pair. Examples of the augmentations used in this study are shown in Fig.~\ref{fig:augmentations}.
After completing the augmentation process, we trained our Siamese network for 10 epochs with a learning rate of 1e-3.

\subsection{Frame sampling from videos}\label{sec:downsampling}

Comparing large datasets of videos in a frame-wise manner presents significant computational overhead. It is not uncommon for footage to be published at 60 frames per second for modern quality standards, which under the quadratic scaling of pairwise sampling would lead to 3,600 comparisons between just two seconds of footage. It is therefore necessary to define a standard for subsampling the frames in a given video to reduce the number of comparisons that need to be made, while maintaining a sufficient level of accuracy in our comparisons. This can be done by careful application of our Siamese network.

In practical terms, we sequentially select frames from a given video such that each frame is considered dissimilar by the model relative to all previously extracted frames. Hereafter, we refer to this sampled set as ``representative frames'' since they are a subset of frames similar to all frames in the video. This process requires setting a threshold for similarity in frame comparisons. During a period of testing considering the number of frames returned and processing time, we found that 0.2 was appropriate. Using this threshold, we were able to reduce the average number of frames considered from each video to fewer than 10\% of the total, reducing the average number of comparisons required between two videos to 1\% of the total framewise comparisons. This application has advantages over the sampling of frames at regular intervals in that it ensures that all visually distinct frames are identified, regardless of the duration of such imagery. The process for identifying representative frames is explained in more detail in Algorithm~\ref{alg:extract frames}.

\begin{algorithm}
\caption{Algorithm for extracting representative frames}
\label{alg:extract frames}
\begin{algorithmic}[1]
    \Require $F$, a list of video frames
    \Require $sim$, an image similarity function
    \Require $s \in \mathbb{R}$, a similarity threshold
    \State Initialise $R$, an empty list of representative frames
    \While{$F$ is not empty}
    \State Add the first frame $f \in F$ to $R$
    \State Calculate the similarity of $f$ to all frames $g \in F \setminus f$
    \State Remove frames $g$ from $F$ where $sim(f,g) \leq s$
    \EndWhile
    \State \Return List $R$ of representative frames
\end{algorithmic}
\end{algorithm}

\subsection{Identifying reused footage}\label{sec:identifying_reused_footage}

With these parts in place, we are now able to describe the steps required to apply this method to a set of videos. The first step is to use Algorithm~\ref{alg:extract frames} to preprocess the videos in the reference set and prepare a database of their representative frames. This process of representative frame extraction can be done ahead of time and stored for efficient application to new videos as they are added to the dataset. 

Given a new video to be compared, its set of representative frames are extracted in the same manner, before the comparison process proceeds as detailed in Algorithm~\ref{alg:compare frames}. The representative frames for each video are compared in turn, and should one or more pair of frames be matched for a given pair of videos under a chosen similarity threshold, we record only the closest match. As with the representative frame extraction, we found again that a threshold of 0.2 was appropriate for video matching after some testing.

\begin{algorithm}
    \caption{Algorithm for comparing representative frames}
    \label{alg:compare frames}
    \begin{algorithmic}[1]
    \Require $R_1$, a set of sets $V_i$ of representative frames for all videos in the reference dataset
    \Require $R_2$, a set of representative frames of a new comparison video
    \Require $sim$, an image similarity function
    \Require $s \in \mathbb{R}$, a similarity threshold
    \State Initialise $M$, an empty list of matched frames
    \For{$f \in R_2$}
    \For{$V_i \in R_1$}
    \State Initialise $C$, an empty list of candidate matches
    \State Initialise $S$, an empty list of candidate similarities
    \For{$f_i \in V_i$}
    \If{$sim(f,f_i) \leq s$}
    \State Add $(V_i,f_i)$ to $C$, $sim(f,f_i)$ to $S$
    \EndIf
    \EndFor
    \State Add $(V_i,f_i) \in C$ to $M$ such that $sim(f,f_i)=min(S)$
    \EndFor
    \EndFor
    \State \Return $M$
    \end{algorithmic}
\end{algorithm}

\subsection{Model evaluation}

We define a test case for our model that requires it to identify which footage has been included in a longer video comprising many extracts from stock footage. We will use this terminology throughout the rest of the manuscript: ``clips'' refers to short stock footage videos collected from Pixabay, whereas ``composites'' refers to any video featuring several clips edited together.

We task our model with identifying the constituent clips of a series of composites from a set of all clips. Each composite is made up of six clips sampled from the full set of 125 clips, and a random 10-second span is sampled from each clip. These clips are concatenated with an additional transition frame added between clips. We do this 125 times to produce an evaluation set of 125 composite videos. 

\section{Results}\label{sec:results}
This section describes the overall evaluation process and the performance of the proposed model in identifying constituent clips within longer composite videos. We first compare the performance of different backbone architectures using precision, recall, and balanced accuracy as metrics across a range of similarity thresholds to determine the optimal backbone and similarity threshold. We then investigate the characteristics of video footage that contribute to retrieval errors by analysing the frequency with which each Pixabay video is incorrectly classified as either a false positive or a false negative. Finally, we evaluate the retrieval capability of the proposed method using the recall@($k$) metric, where $k = 1, 6, 12$, to assess the extent to which constituent clips are successfully retrieved among the top-$k$ most similar videos.

\begin{figure}[ht]
    \centering
    \begin{subfigure}{0.80\textwidth}
        \includegraphics[width=\linewidth]{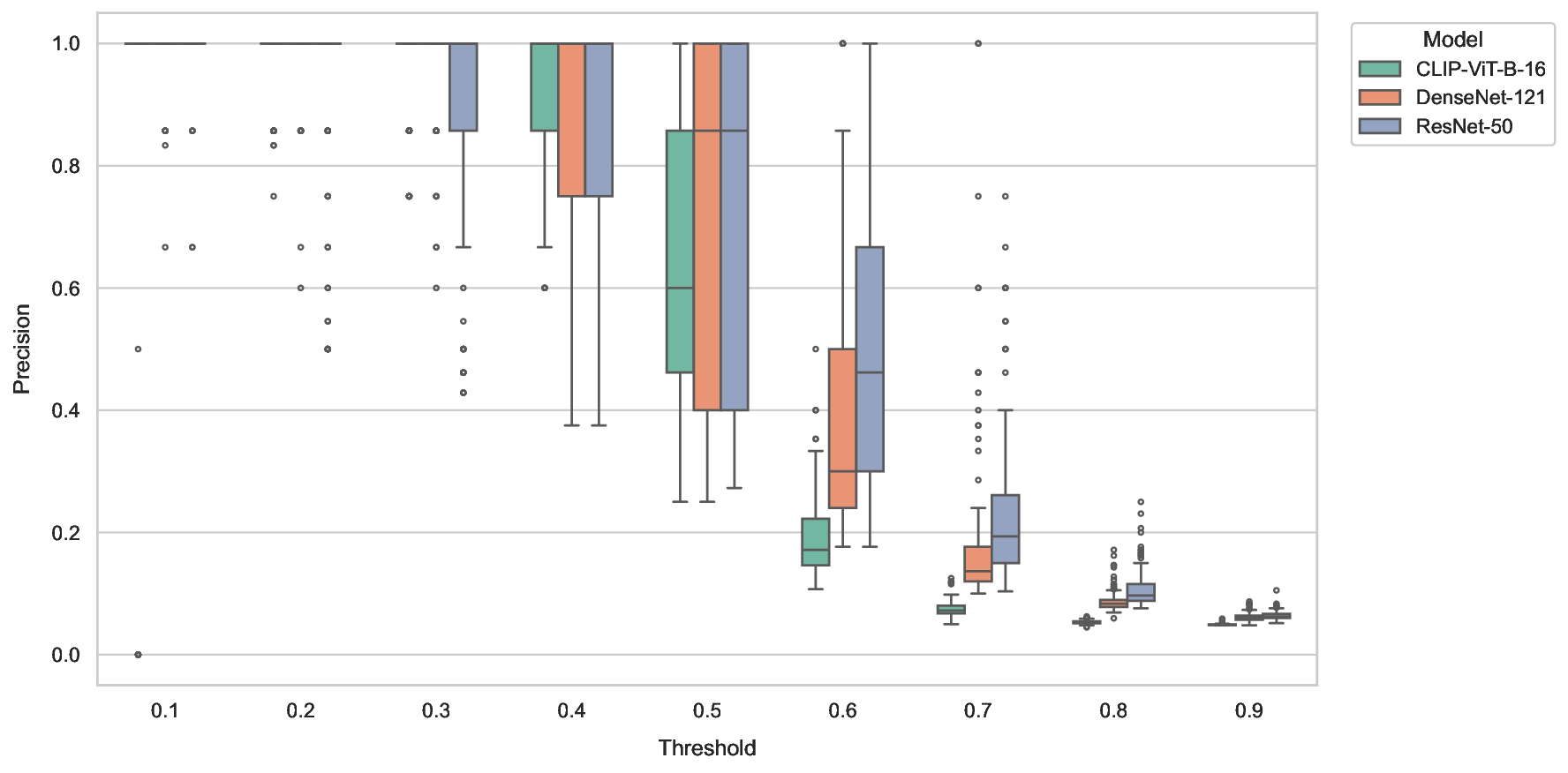}
        \caption{Precision}
        \label{fig:precision}
    \end{subfigure} \\
    \begin{subfigure}{0.80\textwidth}
        \includegraphics[width=\linewidth]{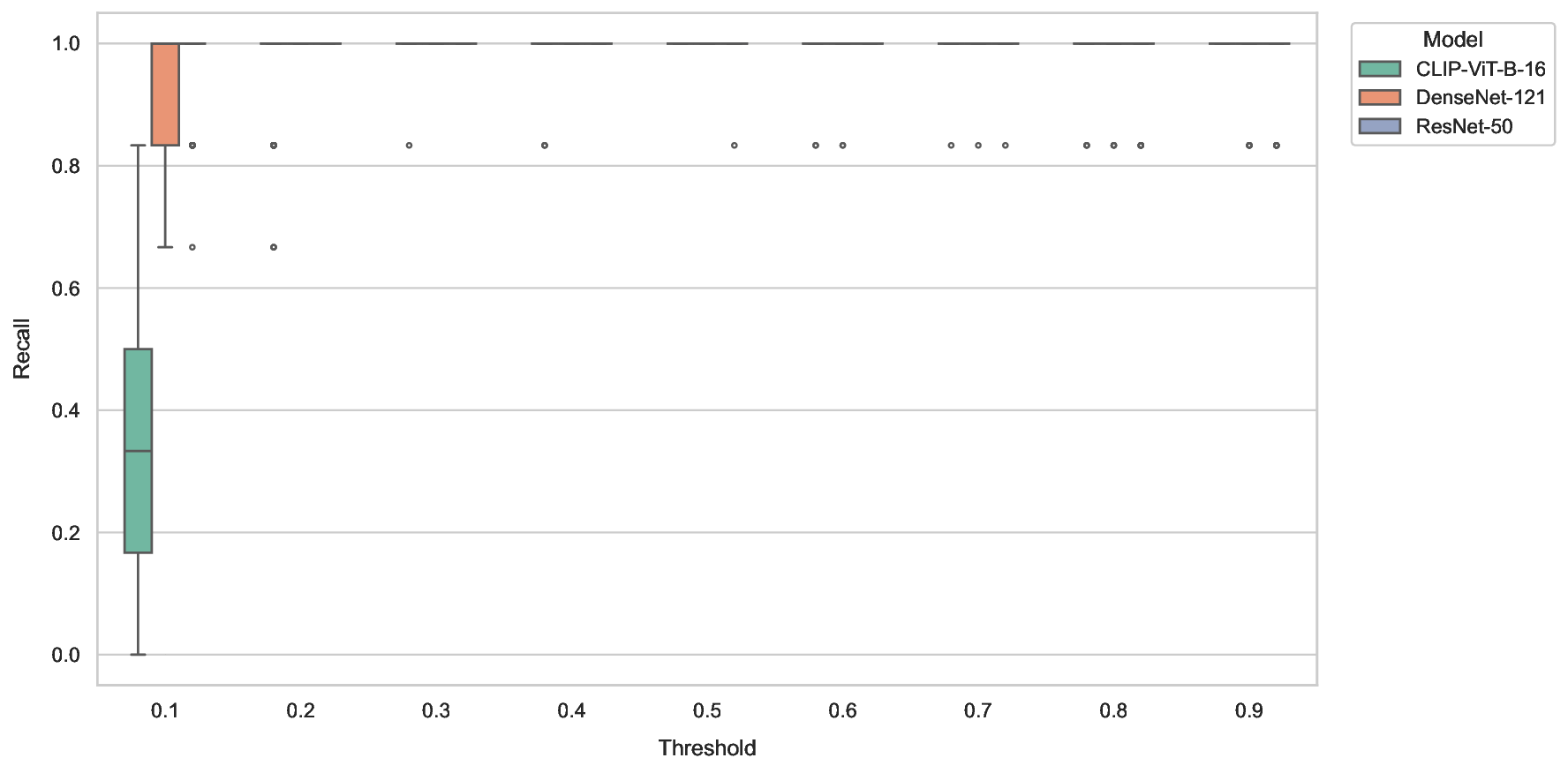}
        \caption{Recall}
        \label{fig:recall}
    \end{subfigure} \\
    \begin{subfigure}{0.80\textwidth}
        \includegraphics[width=\linewidth]{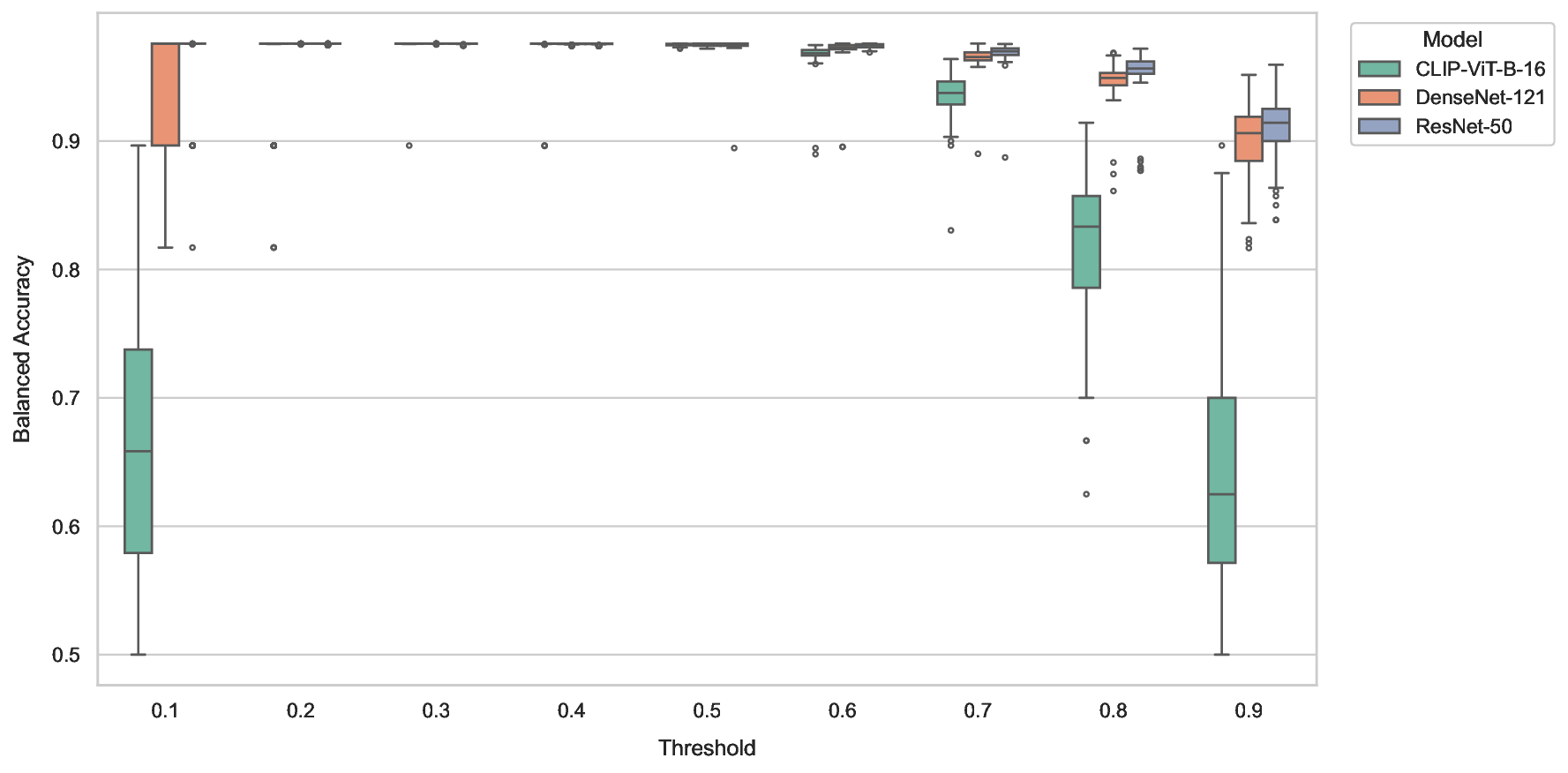}
        \caption{Balanced accuracy}
        \label{fig:bal_accuracy}
    \end{subfigure}
    \caption{Performance statistics for our method when applied to the synthetic videos using the DenseNet-121, ResNet-50 and Clip-ViT-Base-Patch-16 backbones. The method displays a preference for recall over precision, and lower precision is the cause for lower balanced accuracy in most cases.}
    \label{fig:performance stats}
\end{figure}

 Fig.~\ref{fig:performance stats} presents the precision, recall, and balanced accuracy achieved by the proposed framework using ResNet-50, DenseNet-121, and CLIP backbones across a range of similarity thresholds. All three backbones typically achieve perfect recall for similarity thresholds greater than 0.1, indicating that all constituent clips are successfully identified. Precision and balanced accuracy are highest at similarity thresholds between 0.2 and 0.5, however, both metrics progressively decline beyond this range. The model trained on a CLIP backbone exhibits the largest degradation, reaching values as low as 0 for precision and 0.5 for balanced accuracy. These results indicate that increasing the similarity threshold allows larger distances between frame embeddings to be considered as matches, resulting in an increased number of false-positive detections. Overall, each backbone performs well with a similarity threshold of 0.2 or 0.3, with minor trade-offs between precision, recall, and balanced accuracy.



\begin{figure}[!h]
    \centering
    \includegraphics[width=\linewidth]{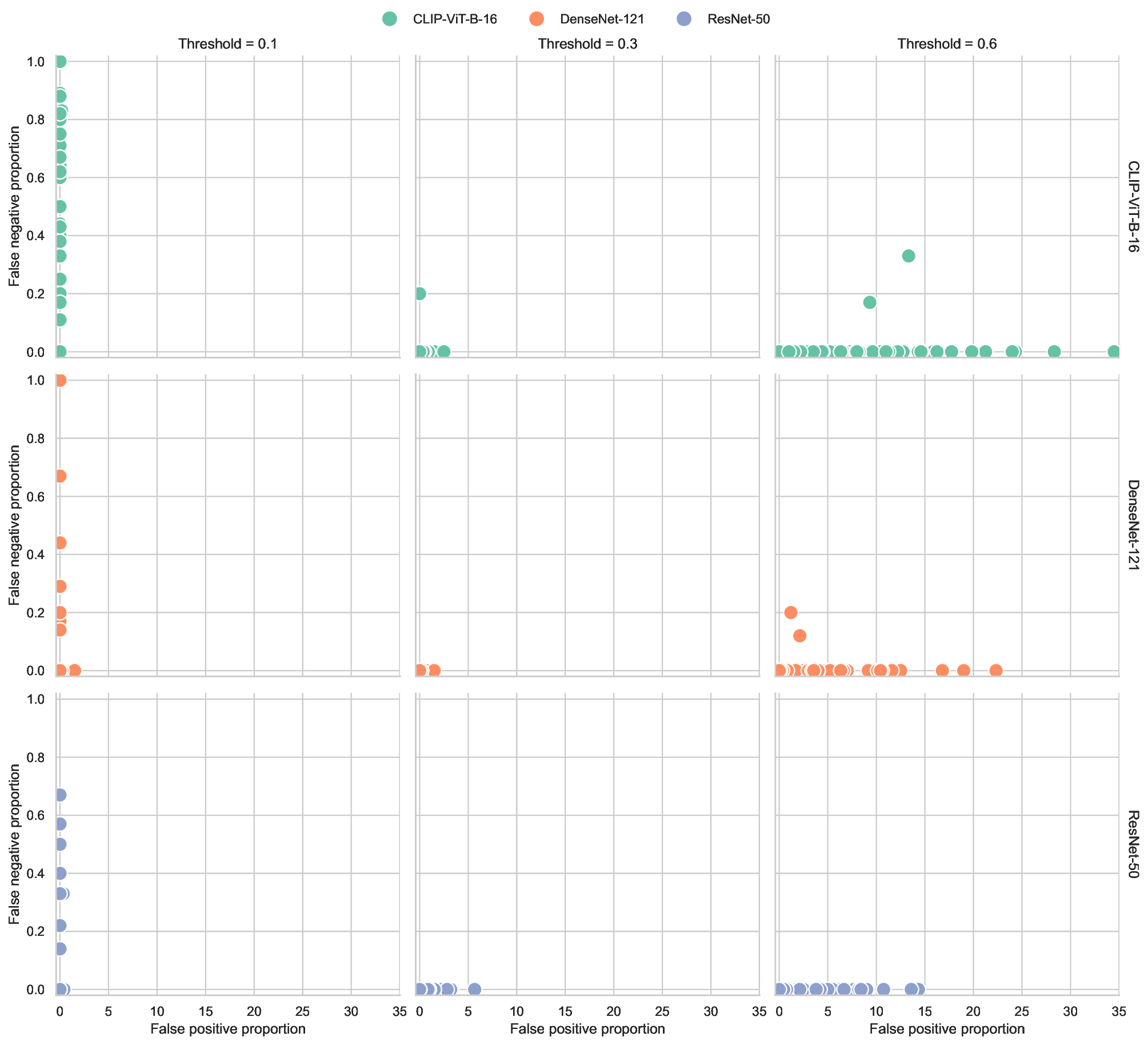}
    \caption{False positive and false negative proportions for each Pixabay video in the dataset, normalised by the number of composite videos they appear in. Some videos are shown to be hard to detect (false negative rate of 1) or commonly detected incorrectly (false positive rate $> 5$).}
    \label{fig:fp-vs-fn}
\end{figure}

Our next avenue of investigation considers whether there are any patterns in video footage that cause issues with our method. To do this, we consider the frequency with which each of the Pixabay videos was labelled as either a false positive or a false negative in our dataset. Fig.~\ref{fig:fp-vs-fn} shows this for each video, normalised by the number of composite videos it appears in (i.e. a false positive proportion of 10 means that a video is incorrectly identified 10 times for each time it was included in a composite video). 
As the similarity threshold increases, the proportion of false positives rises, while the proportion of false negatives decreases. This behaviour is expected, as a larger threshold allows a greater number of frame pairs to be classified as matches. This overall trend reflects the role of the threshold. Smaller thresholds require a greater degree of similarity between frames, therefore biasing towards precision at the expense of recall. Higher thresholds are more permissive, therefore leading to more potential matches to be identified.

With a threshold of 0.3 identified as near optimal in Fig.~\ref{fig:performance stats}, all backbones show a good balance between false-positive and false-negative proportions, with the majority of videos clustered near the origin, indicating low error rates. The CLIP backbone is the only one to see non-zero false-negative proportion, and ResNet-50 sees slightly more false positives than the other backbones. 

\begin{figure}[!ht]
    \centering
    \begin{subfigure}{0.92\textwidth}
        \includegraphics[width=\linewidth]{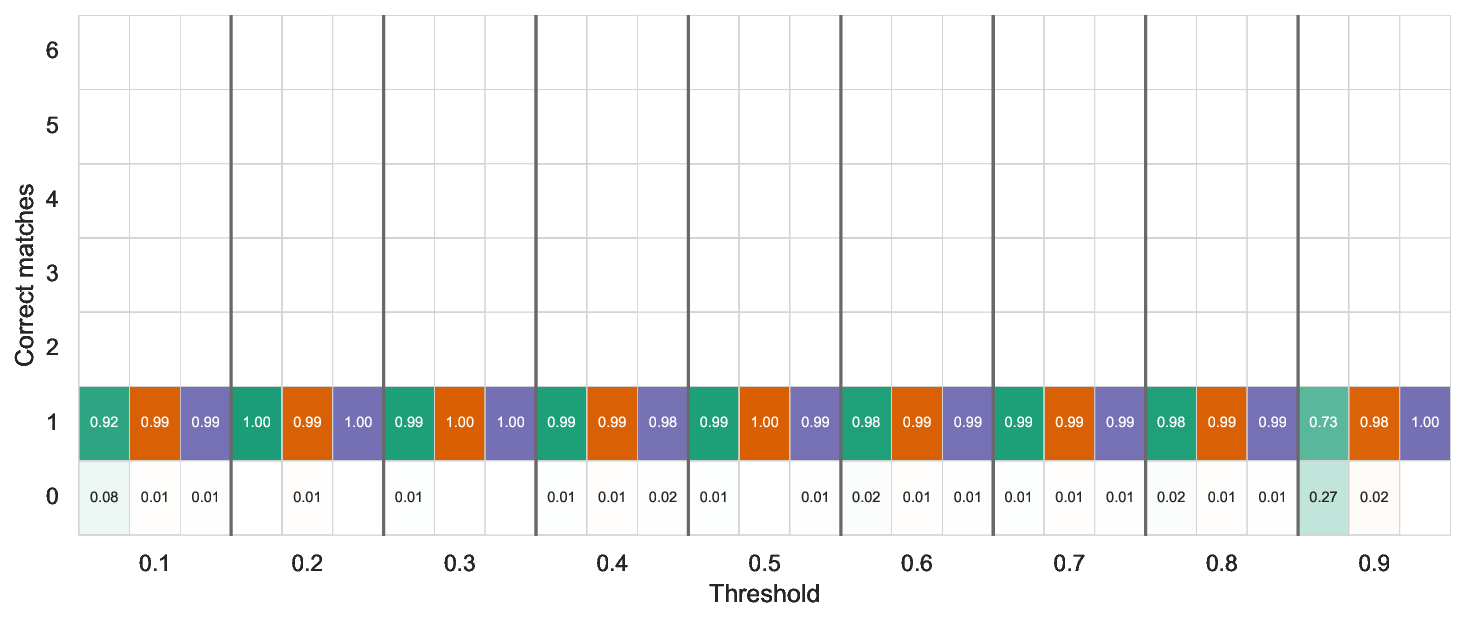}
        \caption{$k=1$}
        \label{fig:prop_recall_k1}
    \end{subfigure} \\
    \begin{subfigure}{0.92\textwidth}
        \includegraphics[width=\linewidth]{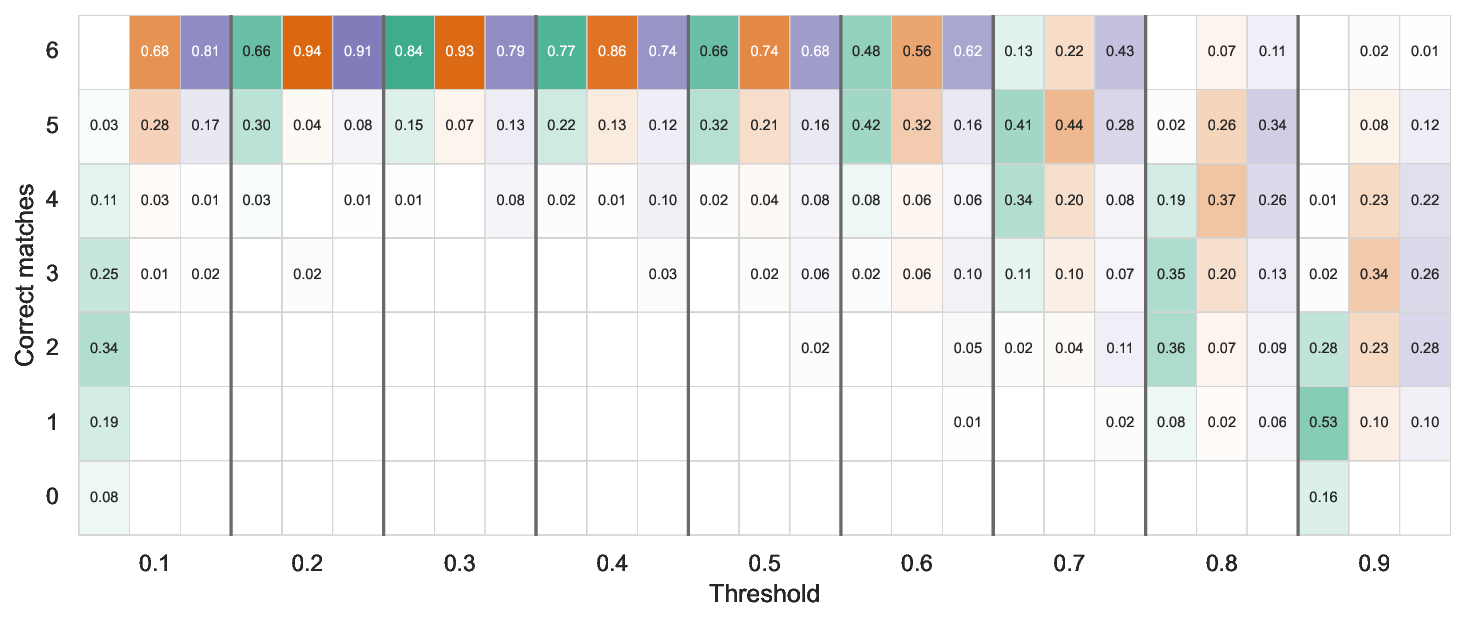}
        \caption{$k=6$}
        \label{fig:prop_recall_k6}
    \end{subfigure} \\
    \begin{subfigure}{0.92\textwidth}
        \includegraphics[width=\linewidth]{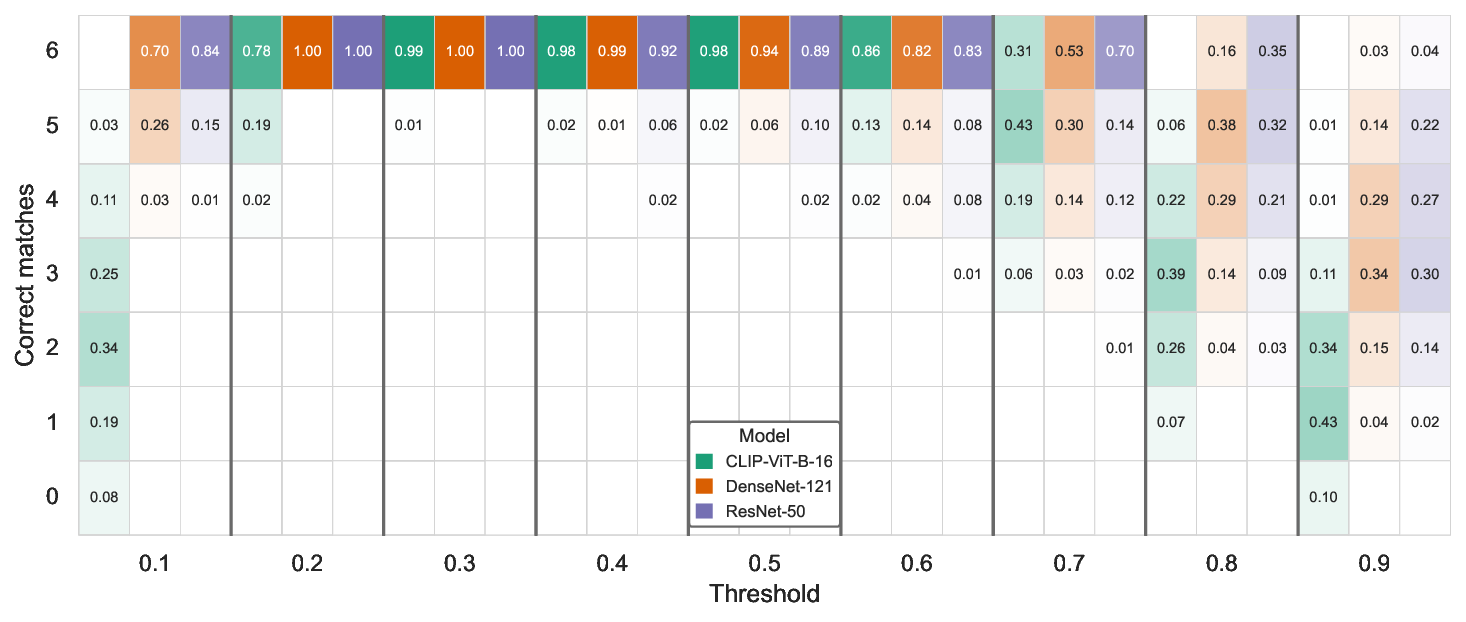}
        \caption{$k=12$}
        \label{fig:prop_recall_k12}
    \end{subfigure}
    \caption{Recall@($k$) performance for $k=1,6,12$. Each cell illustrates the proportion of composite videos correctly identifying the relevant number of included clips.}
    \label{fig:prop_recall_by_threshold}
\end{figure}

Recall@($k$) has been a popular metric to test the performance of retrieval or embedding models ~\cite{Chen20,vitr2023,blip22023}. Here we calculate recall@($k$) as the extent to which the included clips are detected when we consider the closest one, six and 12 videos. These values reflect the use case here, allowing our model to recommend a single best match, up to one match per included clip and up to twelve matches per included clip.  Fig.~\ref{fig:prop_recall_by_threshold} presents the recall@($k$) results across a range of similarity thresholds. A few patterns can be seen across all backbone architectures and thresholds. The recall@(1) results indicate that, for the majority of composite videos, the highest-ranked retrieved video corresponds to one of the six constituent clips. Recall@(6) results indicate that all model configurations successfully recover most of the six constituent clips for most composite videos by the better performing thresholds of 0.2 and 0.3 (and typically perform much better). Finally, recall@(12) results demonstrate that the proposed framework is consistently able to retrieve all six constituent clips when the top twelve most similar videos are considered. Performance once again degrades at low thresholds (0.1) and higher thresholds (>0.5). 



These results give a clear inference for the performance of this method. Firstly, identifying the ``easiest'' match in a given composite video is managed well by the model (i.e. the recall at 1 performance). This performance drops off as we attempt to retrieve all clips within the top-6 matches and reaches perfect recall as we increase the number of matches to 12. This suggests that given an appropriate number of selections, our method can achieve excellent recall. 

\begin{figure}[!ht]
    \centering
        \includegraphics[width=\linewidth]{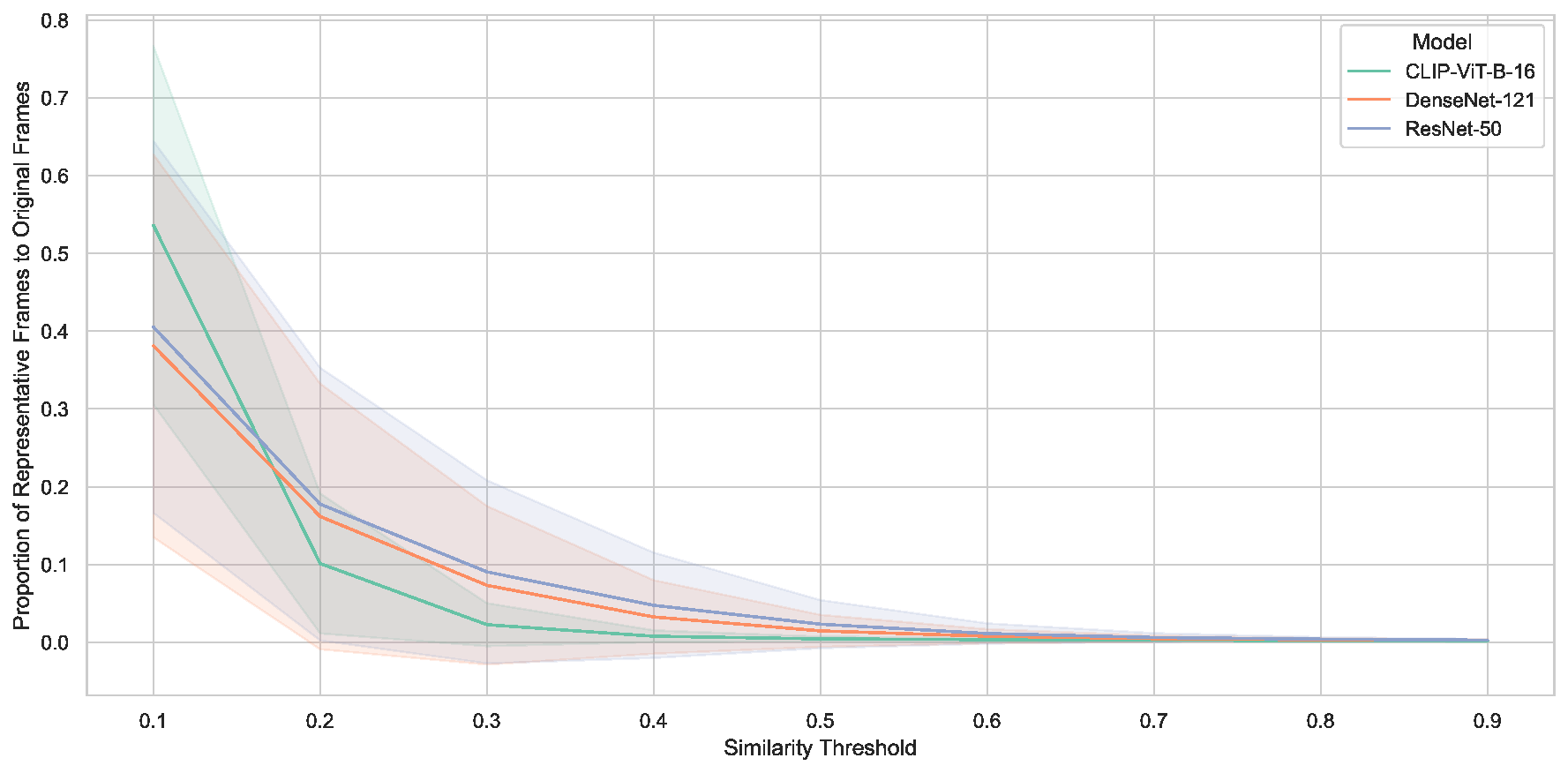}
    \caption{Mean proportion of all frames identified as representative frames at different thresholds over all videos (clips and composites). Shaded regions indicate one standard deviation from the mean.}
    \label{fig:downsampling}
\end{figure}

Our final result discusses the utility of our methodology as an efficient means of downsampling frames from a given video to a subset containing each of the distinct visual elements. In Fig.~\ref{fig:downsampling} we show the relative size of the representative frame sets identified in a random sample of our video data at a range of different similarity thresholds. A few patterns emerge from this experiment. Firstly, application of this technique with even modest thresholds sees notable reductions in the number of frames retained as shown in Fig.~\ref{fig:downsampling}. A threshold of 0.1 sees a 45-60\% reduction in frames on average under all model backbones. This improves as the threshold increases to a greater than 99\% reduction at a threshold of 0.6. In the region of thresholds displaying better performance under the other metrics, we see average reductions of over 90\% of the original frames. 

\section{Discussion}\label{sec:discussion}

In this paper we introduced a novel method of identifying video subsequences in longer footage using contrastive learning. We proposed three different Siamese networks with different encoder backbones. We outlined how these models can be used to extract representative frames, that is a subset of the frames such that all frames in the full video are visually similar to at least one frame in the subset. Finally, we tested these models on a series of composite videos combining known excerpts from a set of stock footage collected from Pixabay.

Our experiments found that the different backbones presented trade offs for each of the criteria we tested, but typically all performed well for similarity thresholds of 0.2 and 0.3. Selection between the backbones should be determined by the balance between the reduction in the size of the representative frame sets (and subsequent speed improvements), precision and recall over different similarity thresholds. Choosing DenseNet-121 or ResNet-50 and a threshold between 0.2 and 0.3 will work in most instances, although a CLIP-ViT-B-16 backbone performs better should the threshold need to be increased to 0.4. 

When testing on the synthetic data, we found that this model displayed excellent recall@($k$) on average. This preference for recall is something that we consider desirable in a potential application of this method: where a content producer has a large body of their own work and seeks an indication of whether any of it appears in another piece of footage before manually reviewing the results. In this instance, a preference for recall will minimise the chance of missed videos in such a dataset. 

We also note that our test case is somewhat biased against precision by the nature of the synthetic videos and the sets of videos matched. By construction, we know that each synthetic video contains exactly six videos. The set of videos recovered by our method on the other hand is only bounded by the number of representative frames identified in the comparator video (since we identify the best match to each). In practice, this typically leads to a larger number of matches with a given video, and therefore significantly hinders the ability for precision to approach one. We expect this to be a quirk of stock footage rather than a universal issue with the method, and is one that should be alleviated by the typically greater visual complexity of video footage in general.

We saw in Fig.~\ref{fig:downsampling} that our method of downsampling frames was quickly able to reduce the total number of frames to be considered to less than 10\% of the original number on average. This reduction has significant impact on the running time of comparison methods that scale superlinearly with the input data (as is the case with our reuse detection method). This observation has implications for other efforts for content analysis of videos. The simplest approach is to sample frames periodically (e.g.~\cite{Yoon_2023_CVPR,Vedula2017,larocca2025evaluatingaicapabilitiesdetecting}) or based on features such as scene transitions (e.g.~\cite{Hannane2016,Tang_2025_CVPR}) to collect a sample with sufficient coverage to reflect all themes included in the footage. While this may work in general, there are extreme cases in which it can oversample (e.g. an extended scene without much visual change) or undersample (e.g. fast moving footage with many cuts in short succession). In such cases, our method is preferable as it will identify a sample of visually distinct frames based on the threshold chosen.

In terms of the threshold that should be applied, our testing shows that a value of 0.2 or 0.3 leads to excellent performance on our test dataset. While this was the optimal value in our testing, Fig.~\ref{fig:downsampling} indicates that a range of threshold values can have a similar impact on the total number of representative frames. As a result, there may be value in tuning of the threshold for other datasets depending on the purpose to be achieved and dataset characteristics. A larger threshold will lead to fewer representative frames (and therefore faster comparisons), but a worse performance. Reducing the threshold will invert this pattern. The most important factor to consider when adjusting the threshold in practice is the relationship between the one used for representative frame identification and the one used for video matching. We chose to both the same to maximise the precision (as increasing the matching threshold will lead to more matches being identified). In any case, should the matching threshold be less than or equal to the representative frame threshold there will be a decline in the recall of the method as the representative frame set may no longer be matched with all frames in the original video.

During our testing of our methodology, we identified several cases that were particularly challenging for the contrastive learning approach to recognise. The first such case was dark footage, although this is not surprising since such footage necessarily contains less visual information. In additional experiments on real-world examples (not reported here), we found that heavily edited footage was challenging for our model to detect. The same content published with different aspect ratios was difficult to recover as these sometimes coincided with alterations to the placement of elements within a frame. Picture-in-picture inclusion caused similar issues. We leave it to future work to investigate whether a patch-based approach would alleviate these two challenges.

The method and application we detail here are a demonstration that Siamese networks trained using contrastive learning are an effective tool for identifying video content reuse. Future extensions of this work should monitor the development of additional models that can be incorporated as backbones, or the development of new tools for efficient retrieval of similar frames from a database. In addition, there is scope for application of this method to a dataset of real-world videos (assuming suitable ground-truth data can be collated). While such a task would be an additional challenge, such insights would provide a valuable compliment to our discussion here of a best-case scenario. We also note the growing capabilities of multimodal models, especially the possibility of tasking such models with comparing two videos directly or incorporating their audio tracks.

In this manuscript we have shown that Siamese networks produce effective pipelines for video content reuse detection. The growing number of vision models and their capabilities presents a ripe opportunity for refining new methods for video recognition. We hope that this work is the first of many in this direction.

\bibliography{Ref}
\bibliographystyle{ieeetr}

\newpage

\makeatletter 
\renewcommand{\thefigure}{S\@arabic\c@figure}
\renewcommand{\thetable}{S\@arabic\c@table}
\makeatother

\setcounter{figure}{0}
\setcounter{table}{0}

\begin{appendix}
\section{Supplementary data}

\begin{table}[!h]
    \centering
    \begin{tabular}{|c|c|} \hline
       Video URL  & Search term \\ \hline
        https://pixabay.com/videos/people-walking-sun-sky-traffic-197531/ & people \\ \hline
https://pixabay.com/videos/human-person-walk-the-crowd-women-95193/ & people\\ \hline
https://pixabay.com/videos/boys-girls-group-friends-young-51337/ & people \\ \hline
https://pixabay.com/videos/hug-friends-talk-nature-people-159243/ & people \\ \hline
https://pixabay.com/videos/dance-music-to-dance-club-disco-1670/ & people \\ \hline
https://pixabay.com/videos/charity-village-people-help-ethnic-139400/ & people \\ \hline
https://pixabay.com/videos/lunch-meal-people-tea-cup-tea-2339/ & people \\ \hline
https://pixabay.com/videos/people-commerce-shop-busy-mall-6387/ & people \\ \hline
https://pixabay.com/videos/shoes-walking-feet-legs-many-3627/ & people \\ \hline
https://pixabay.com/videos/new-york-city-manhattan-people-cars-1044/ & people \\ \hline
https://pixabay.com/videos/monaco-street-city-people-car-31661/ & people \\ \hline
https://pixabay.com/videos/photographer-beach-photography-91/ & people \\ \hline
https://pixabay.com/videos/relaxing-sitting-lake-shore-1049/ & people \\ \hline
https://pixabay.com/videos/puzzle-puzzling-game-play-playing-132/ & people \\ \hline
https://pixabay.com/videos/concert-light-effects-cheering-1630/ & people \\ \hline
https://pixabay.com/videos/india-people-asia-street-asian-1643/ & people \\ \hline
https://pixabay.com/videos/homework-girl-student-laptop-study-1028/ & people \\ \hline
https://pixabay.com/videos/airport-people-crowd-busy-36510/ & people \\ \hline
https://pixabay.com/videos/walking-busy-people-pace-step-21782/ & people \\ \hline
https://pixabay.com/videos/seoul-people-night-city-23258/ & people \\ \hline
https://pixabay.com/videos/path-people-helicopter-mountains-44702/ & people \\ \hline
https://pixabay.com/videos/people-sunset-stroll-moscow-glare-84973/ & people \\ \hline
https://pixabay.com/videos/summer-beach-sea-ocean-water-24541/ & people \\ \hline
https://pixabay.com/videos/walk-couple-romantic-sky-romance-20454/ & people \\ \hline
https://pixabay.com/videos/winter-snow-people-walk-to-stroll-188778/ & people \\ \hline
https://pixabay.com/videos/new-york-city-nigh-brooklyn-bridge-7260/ & city \\ \hline
https://pixabay.com/videos/city-night-lights-urban-cityscape-3134/ & city \\ \hline
https://pixabay.com/videos/landscape-nature-sky-mountains-24746/ & city \\ \hline
https://pixabay.com/videos/rain-raindrop-wet-water-window-28236/ & city \\ \hline
https://pixabay.com/videos/river-urban-city-architecture-44509/ & city \\ \hline
https://pixabay.com/videos/night-city-city-lights-lights-28860/ & city \\ \hline
https://pixabay.com/videos/dubai-skyline-architecture-31956/ & city \\ \hline
https://pixabay.com/videos/upper-palatinate-bavaria-landscape-31556/ & city \\ \hline
https://pixabay.com/videos/sunset-nature-sky-dubai-111204/ & city \\ \hline
https://pixabay.com/videos/castle-church-tower-city-29224/ & city \\ \hline
https://pixabay.com/videos/alone-person-man-people-guitar-46637/ & city \\ \hline
https://pixabay.com/videos/skyscrapers-sunset-skyline-clouds-80724/ & city \\ \hline
https://pixabay.com/videos/subway-metro-train-urban-city-1003/ & city \\ \hline
https://pixabay.com/videos/intersection-cars-pedestrian-night-180386/ & city \\ \hline
https://pixabay.com/videos/seoul-city-streets-road-night-view-21985/ & city \\ \hline
https://pixabay.com/videos/skyscrapers-buildings-city-urban-91744/ & city \\ \hline
https://pixabay.com/videos/ocean-beach-waves-breaking-waves-160767/ & city \\ \hline
https://pixabay.com/videos/cars-roundabout-rondo-flowers-road-168811/ & city \\ \hline
https://pixabay.com/videos/sunset-canal-city-light-afternoon-230248/ & city \\ \hline
https://pixabay.com/videos/city-sunset-buildings-skyscrapers-173137/ & city \\ \hline
https://pixabay.com/videos/london-eye-london-england-landmark-202368/ & city \\ \hline
https://pixabay.com/videos/lake-fireworks-night-view-city-225661/ & city \\ \hline
https://pixabay.com/videos/basketball-hoop-net-urban-street-208442/ & city \\ \hline
https://pixabay.com/videos/river-city-bridge-architecture-188057/ & city \\ \hline
https://pixabay.com/videos/building-castle-japanese-castle-250395/ & city \\ \hline

    \end{tabular}
    \caption{Video URLs for ``people'' and ``city'' footage collected from Pixabay in February 2025.}
    \label{tab:pixabay urls}
\end{table}

\begin{table}[h]
    \centering
    \begin{tabular}{|c|c|} \hline
       Video URL  & Search term \\ \hline
https://pixabay.com/videos/trees-forest-autumn-scenic-4k-185726/ & landscape \\ \hline
https://pixabay.com/videos/beach-sea-ocean-sand-drone-aerial-186115/ & landscape \\ \hline
https://pixabay.com/videos/ocean-waves-rocks-sand-coast-213026/ & landscape \\ \hline
https://pixabay.com/videos/beach-coastline-travel-tourism-161045/ & landscape \\ \hline
https://pixabay.com/videos/ocean-beach-sunset-waves-dawn-183960/ & landscape \\ \hline
https://pixabay.com/videos/bird-seagull-boat-ocean-sea-223461/ & landscape \\ \hline
https://pixabay.com/videos/ocean-sea-waves-rocks-pacific-217115/ & landscape \\ \hline
https://pixabay.com/videos/drone-drones-drone-images-nature-216058/ & landscape \\ \hline
https://pixabay.com/videos/waterfall-mountain-stream-flow-228847/ & landscape \\ \hline
https://pixabay.com/videos/gorge-quartzite-rock-aerial-descent-232185/ & landscape \\ \hline
https://pixabay.com/videos/tractor-field-wheat-harvest-232561/ & landscape \\ \hline
https://pixabay.com/videos/bridge-wooden-bridge-197564/ & landscape \\ \hline
https://pixabay.com/videos/leaves-water-river-stream-autumn-240320/ & landscape \\ \hline
https://pixabay.com/videos/temple-pond-garden-kinkakuji-temple-214940/ & landscape \\ \hline
https://pixabay.com/videos/ship-vessel-sea-boat-ocean-water-237249/ & landscape \\ \hline
https://pixabay.com/videos/ocean-sea-beach-sunset-sand-204565/ & landscape \\ \hline
https://pixabay.com/videos/mountains-mount-fuji-volcano-winter-234930/ & landscape \\ \hline
https://pixabay.com/videos/mountain-himalayas-himachal-manali-199958/ & landscape \\ \hline
https://pixabay.com/videos/pagoda-mountains-japan-japanese-240841/ & landscape \\ \hline
https://pixabay.com/videos/sunset-mountains-clouds-tyrol-88207/ & landscape \\ \hline
https://pixabay.com/videos/path-bridge-forest-lake-water-241802/ & landscape \\ \hline
https://pixabay.com/videos/drone-landscape-green-greenland-236893/ & landscape \\ \hline
https://pixabay.com/videos/mount-fuji-morning-highway-202982/ & landscape \\ \hline
https://pixabay.com/videos/road-cars-dwingeloo-netherlands-240330/ & landscape \\ \hline
https://pixabay.com/videos/autumn-fall-leaves-foliage-trees-242945/ & landscape \\ \hline
https://pixabay.com/videos/fish-to-swim-animal-aquarium-156111/ & climate \\ \hline
https://pixabay.com/videos/sky-sea-ocean-waves-view-nature-155041/ & climate \\ \hline
https://pixabay.com/videos/factory-power-plant-industry-smoke-79911/ & climate \\ \hline
https://pixabay.com/videos/ocean-sea-wave-cliff-coast-beach-156323/ & climate \\ \hline
https://pixabay.com/videos/reef-waves-sea-coral-beach-ocean-153601/ & climate \\ \hline
https://pixabay.com/videos/waterfall-river-park-pond-sunny-258399/ & climate \\ \hline
https://pixabay.com/videos/mountains-fog-cloudy-top-high-251451/ & climate \\ \hline
https://pixabay.com/videos/river-waterfall-nature-mountain-259672/ & climate \\ \hline
https://pixabay.com/videos/house-weather-rain-rainfall-127850/ & climate \\ \hline
https://pixabay.com/videos/flower-yellow-sea-summer-nature-171219/ & climate \\ \hline
https://pixabay.com/videos/piranha-fish-tooth-beast-hunter-156143/ & climate \\ \hline
https://pixabay.com/videos/waterfall-cascade-river-that-120256/ & climate \\ \hline
https://pixabay.com/videos/forest-woods-deforestation-230496/ & climate \\ \hline
https://pixabay.com/videos/waterfall-beautiful-river-park-259366/ & climate \\ \hline
https://pixabay.com/videos/grass-herbs-leaves-garden-wind-215157/ & climate \\ \hline
https://pixabay.com/videos/river-waterfall-flow-cascade-waves-158405/ & climate \\ \hline
https://pixabay.com/videos/river-waterfall-nature-mountain-259904/ & climate \\ \hline
https://pixabay.com/videos/clouds-cumulus-sky-fluffy-outdoors-64767/ & climate \\ \hline
https://pixabay.com/videos/waterfall-river-park-pond-sunny-259413/ & climate \\ \hline
https://pixabay.com/videos/ocean-sea-light-paradise-god-sun-141367/ & climate \\ \hline
https://pixabay.com/videos/sea-coast-kennedy-cliff-stone-259919/ & climate \\ \hline
https://pixabay.com/videos/snow-melt-water-cold-ice-cream-7951/ & climate \\ \hline
https://pixabay.com/videos/cloud-sky-grey-clouds-cloudy-sky-9151/ & climate \\ \hline
http://pixabay.com/videos/cloud-sky-grey-clouds-cloudy-sky-9153/ & climate \\ \hline
https://pixabay.com/videos/clouds-cumulus-sky-fluffy-outdoors-64759/ & climate \\ \hline

    \end{tabular}
    \caption{Video URLs for ``landscape'' and ``climate'' footage collected from Pixabay in February 2025.}
    \label{tab:pixabay urls 2}
\end{table}

\begin{table}[h]
    \centering
    \begin{tabular}{|c|c|} \hline
       Video URL  & Search term \\ \hline
https://pixabay.com/videos/frog-reeds-green-113403/ & wildlife \\ \hline
https://pixabay.com/videos/mountains-forest-nature-sky-clouds-168787/ & wildlife \\ \hline
https://pixabay.com/videos/sea-ocean-seagulls-birds-sunset-140111/ & wildlife \\ \hline
https://pixabay.com/videos/snail-mollusk-creeping-animal-169471/ & wildlife \\ \hline
https://pixabay.com/videos/flower-bee-anemone-pink-insect-178826/ & wildlife \\ \hline
https://pixabay.com/videos/flower-peony-bee-insect-pollinate-165065/  & wildlife \\ \hline
https://pixabay.com/videos/cow-meadow-pasture-179207/  & wildlife \\ \hline
https://pixabay.com/videos/birds-flock-fly-sky-circle-138891/  & wildlife \\ \hline
https://pixabay.com/videos/chamois-wild-animal-animal-fall-188861/  & wildlife \\ \hline
https://pixabay.com/videos/bees-insects-beehive-honey-bees-178049/  & wildlife \\ \hline
https://pixabay.com/videos/bird-cormorant-beak-plumage-198358/  & wildlife \\ \hline
https://pixabay.com/videos/marsupials-mammal-wildlife-animal-181314/  & wildlife \\ \hline
https://pixabay.com/videos/tiger-predator-dangerous-carnivores-145320/  & wildlife \\ \hline
https://pixabay.com/videos/bird-bee-eater-feathers-colorful-200427/ & wildlife \\ \hline
https://pixabay.com/videos/bird-blue-tit-plumage-feeding-place-187970/ & wildlife \\ \hline
https://pixabay.com/videos/horses-animals-wildlife-running-206294/ & wildlife \\ \hline
https://pixabay.com/videos/birds-red-crowned-cranes-cranes-219862/ & wildlife \\ \hline
https://pixabay.com/videos/sunset-seagulls-birds-gulls-203196/ & wildlife \\ \hline
https://pixabay.com/videos/baboons-apes-cub-family-tree-trunk-216696/ & wildlife \\ \hline
https://pixabay.com/videos/bird-daurian-redstart-beak-feathers-202693/ & wildlife \\ \hline
https://pixabay.com/videos/bird-sperling-house-sparrow-230060/ & wildlife \\ \hline
https://pixabay.com/videos/gull-bird-snow-plumage-sitting-191159/ & wildlife \\ \hline
https://pixabay.com/videos/chameleon-reptile-animal-green-233037/ & wildlife \\ \hline
https://pixabay.com/videos/turtle-ocean-tortoise-wildlife-244754/ & wildlife \\ \hline
https://pixabay.com/videos/fish-underwater-aquarium-swim-110879/ & wildlife \\ \hline

    \end{tabular}
    \caption{Video URLs for ``wildlife'' footage collected from Pixabay in February 2025.}
    \label{tab:pixabay urls 3}
\end{table}

\begin{figure}[h]
    \centering
    \includegraphics[width=\linewidth]{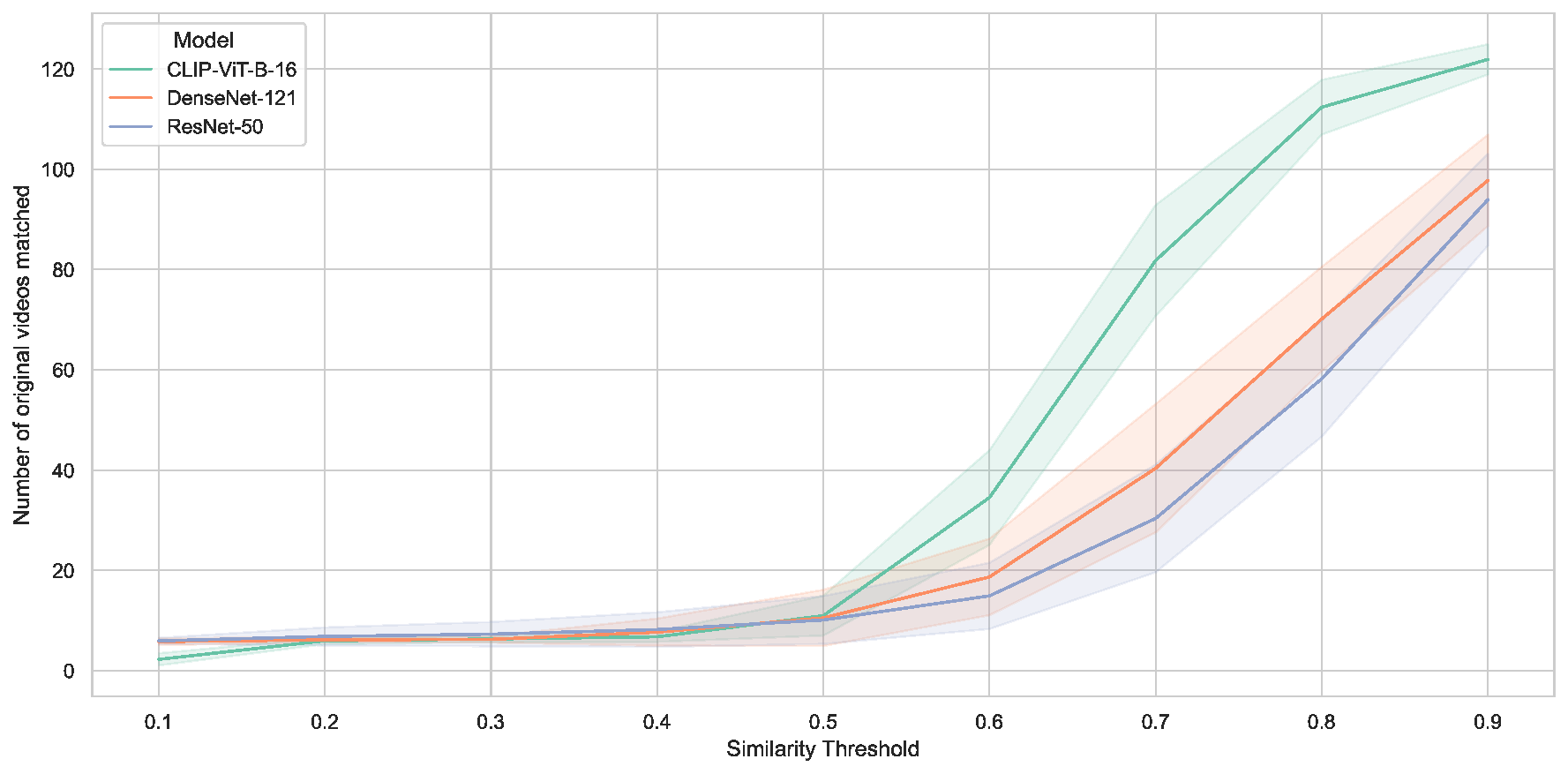}
    \caption{Number of original videos within a given similarity threshold of each composite video. The solid line averages over all composite videos, and the shaded region indicates a range of one standard deviation around the mean.}
    \label{fig:num_matched_by_threshold}
\end{figure}

In Fig.~\ref{fig:num_matched_by_threshold} we investigate the number of original videos that are considered as potential matches under different similarity thresholds. We see here that similarity scores less than 0.5 are typically rare and are on average less than 20\% of all pairwise comparisons. The increase in match quantity increases rapidly until the threshold approaches 0.8, at which point almost all of the original videos are considered potential matches. This suggests that in this case, there is little need to apply a large similarity threshold when focusing on the $k$ closest clips when $k$ is small. The other feature of note in Fig.~\ref{fig:num_matched_by_threshold} is the consistency across each of the composite videos. This suggests that these observed patterns are more representative of typical behaviour and are less likely to be influenced by the particular choice of videos for comparison.

\end{appendix}
\end{document}